\documentclass[10pt,twocolumn,letterpaper]{article}

\usepackage{iccv}
\usepackage{times}
\usepackage{epsfig}
\usepackage{graphicx}
\usepackage{amsmath}
\usepackage{amssymb}
\usepackage{bm}
\usepackage{multirow}
\usepackage{multicol}
\usepackage{makecell}

\usepackage[pagebackref=true,breaklinks=true,letterpaper=true,colorlinks,bookmarks=false]{hyperref}

\iccvfinalcopy % *** Uncomment this line for the final submission

\def\iccvPaperID{4049} % *** Enter the ICCV Paper ID here

\ificcvfinal\fi

\begin{document}

%%%%%%%%% TITLE
\title{DAE-GAN:~Dynamic Aspect-aware GAN for Text-to-Image Synthesis}

\author{
Shulan Ruan\textsuperscript{1}\footnotemark[2]~,
% For a paper whose authors are all at the same institution,
% omit the following lines up until the closing ``}''.
% Additional authors and addresses can be added with ``\and'',
% just like the second author.
% To save space, use either the email address or home page, not both
Yong Zhang\textsuperscript{2}\footnotemark[1]~, 
% \thanks{. }~,
Kun Zhang\textsuperscript{3},
Yanbo Fan\textsuperscript{2},
Fan Tang\textsuperscript{4},
Qi Liu\textsuperscript{1},
Enhong Chen\textsuperscript{1}\footnotemark[1]\\
\textsuperscript{1}\small School of Computer Science and Technology, University of Science and Technology of China\\
\textsuperscript{2}\small Tencent AI Lab,~
\textsuperscript{3}\small Hefei University of Technology,~
\textsuperscript{4}\small Jilin University\\
\small slruan@mail.ustc.edu.cn,
\{zhangyong201303, zhang1028kun, fanyanbo0124, tfan.108\}@gmail.com,
\{qiliuql, cheneh\}@ustc.edu.cn
}

\newcommand{\fullname}{Dynamic Aspect-awarE GAN}
\newcommand{\shortname}{DAE-GAN}

\maketitle
% Remove page # from the first page of camera-ready.
\ificcvfinal\thispagestyle{empty}\fi

\renewcommand{\thefootnote}{\fnsymbol{footnote}}
\footnotetext[2]{Work done during an internship in Tencent AI Lab.}
\footnotetext[1]{Corresponding Authors.}

%%%%%%%%% ABSTRACT
\begin{abstract}
Text-to-image synthesis refers to generating an image from a given text description, the key goal of which lies in photo realism and semantic consistency.
Previous methods usually generate an initial image with sentence embedding and then refine it with fine-grained word embedding.
Despite the significant progress,  
the `aspect' information~(e.g., red eyes) contained in the text, referring to several words rather than a word that depicts `a particular part or feature of something', is often ignored,
which is highly helpful for synthesizing image details. 
How to make better utilization of aspect information in text-to-image synthesis still remains an unresolved challenge.
To address this problem, in this paper, we propose a~\fullname~(\shortname)
that represents text information comprehensively from multiple granularities, including sentence-level, word-level, and aspect-level.
Moreover, inspired by human learning behaviors, we develop a novel Aspect-aware Dynamic Re-drawer~(ADR) for image refinement, in which an Attended Global Refinement~(AGR) module and an Aspect-aware Local Refinement~(ALR) module are alternately employed. 
AGR utilizes word-level embedding to globally enhance the previously generated image,
while ALR dynamically employs aspect-level embedding to refine image details
from a local perspective.
Finally, a corresponding matching loss function is designed to ensure the text-image semantic consistency at different levels.
Extensive experiments on two well-studied and publicly available datasets~(i.e., CUB-200 and COCO) demonstrate the superiority and rationality of our method.
   
\end{abstract}
\section{Introduction}
\label{s:introduction}

% Background
Text-to-image synthesis requires an agent to generate a photo-realistic image according to the given text description.
Due to its significant potential in many applications such as art generation~\cite{zhi2017pixelbrush} and computer-aided design~\cite{chen2018text2shape} but challenging nature,
it is arousing extensive research attention in recent years.

\begin{figure}
	\centering
	\includegraphics[height=48mm, width=79mm]{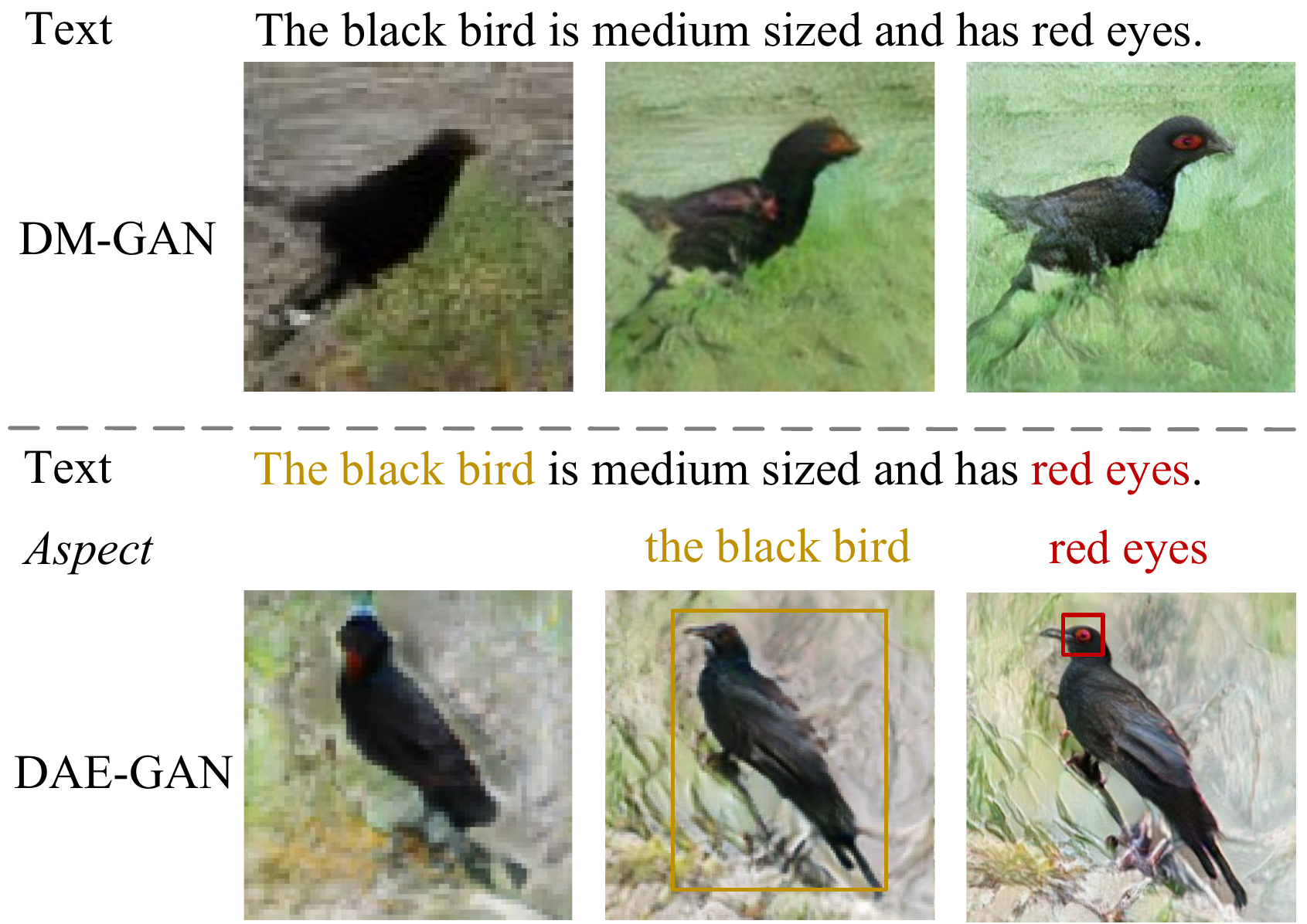}
% 	\vspace{-1mm}
	\caption{Comparisons between DM-GAN~\cite{zhu2019dm} and our~\shortname.
	DM-GAN first generates a low-resolution image with sentence-level information and then refines it with word-level features.
	\shortname~refines images from both global and local perspectives with word-level and aspect information contained.}
	\label{f:examp1e}
 	\vspace{-6mm}
\end{figure}

In the past few years, Generative Adversarial Networks~(GANs)~\cite{goodfellow2014generative} have been proved tremendously successful for this task~\cite{reed2016generative}.
Most existing methods make efforts on the two-stage framework by first generating initial low-resolution images and then refining them to high-resolution ones~\cite{zhang2017stackgan,zhang2018stackgan++,xu2018attngan}.
Among all these methods, the proposal of AttnGAN~\cite{xu2018attngan} plays an extremely important role.
At the initial stage, sentence-level information is employed to generate a low-resolution image. 
Then, at the refinement stage, AttnGAN utilizes word-level features to refine the  previously generated image by
repeatedly adopting attention mechanism to select important words.
Based on AttnGAN, text-to-image synthesis has been pushed by a large step forward~\cite{zhu2019dm,cheng2020rifegan,qiao2019mirrorgan}. 
A synthesis example by DM-GAN~\cite{zhu2019dm} is presented at the top of Figure~\ref{f:examp1e}.

Although remarkable performance has been accomplished with these efforts, there still exist several limitations to be unresolved.
For example, most previous methods only employ sentence-level and word-level features, ignoring the `aspect-level' features.
`Aspect' here refers to several words rather than a word that depicts `a particular part or feature of something'.
There are often multiple aspect terms contained in a sentence to describe an object or a scene from different perspectives,
\textit{e.g.}, `\texttt{the black bird}' and `\texttt{red eyes}' in the text description in Figure~\ref{f:examp1e}. 
Semantic understanding of a sentence is highly dependent on both content and aspect\cite{wang2016attention}.
Both industry and academia have realized the importance of the relationship between aspect term and sentence~\cite{chen2020relation,zheng2020replicate,peng2020knowing}.
In fact, aspect information contained in the text could be helpful for image synthesis, especially for the refinement of local image details.
Though the value of aspect information has been proved, how to make better utilization of aspect information in text-to-image synthesis still remains a big challenge.

Fortunately, some interesting studies of human learning behaviors could give us some inspirations.
Researchers have already demonstrated that human eyes have central vision and peripheral vision~\cite{brandt1973differential,warren1992role,strasburger2011peripheral}. Central vision concentrates on what a person needs at the current time,
while peripheral vision uses observation of the surroundings to support central vision.
Through the dynamic use of central vision and peripheral vision,
we could make an in-depth semantic understanding of text and visual content.

To this end, in this paper,
we propose a novel~\emph{\fullname (\shortname)} for text-to-image synthesis.
To be specific,
we firstly encode text information from multiple granularities comprehensively,
including sentence-level, word-level, and aspect-level.
Then, at the two-stage generation, we first generate a low-resolution image with sentence-level embedding at the initial stage.
Next, at the refinement stage, 
by viewing aspect-level features as central vision and word-level features as peripheral vision,
we develop an \emph{Aspect-aware Dynamic Re-drawer~(ADR)}, which alternately applies an \emph{Attended Global Refinement~(AGR)} module and an \emph{Aspect-aware Local Refinement~(ALR)} module for image refinement.
\emph{AGR} utilizes word-level embedding to globally enhance the previously generated images.
\emph{ALR} dynamically utilizes aspect-level embedding to refine image details from a local perspective.
Finally,  to provide supervision for intermediate synthesis procedures,  a corresponding matching loss function is designed to ensure the text-image semantic consistency.
The bottom of Figure~\ref{f:examp1e} illustrates an example of our proposed method.
When given the aspect `\texttt{the black bird}', our method focuses on adjusting the color of the whole bird body based on the previously generated image. 
When dealing with the aspect `\texttt{red eyes}', our model then correspondingly focuses on the refinement of bird eyes with remaining other parts.

Our main contributions are summarized as follows:
\begin{itemize}
\item We observe the great potential of aspect information and apply it to text-to-image synthesis.
\item We propose a novel~\shortname, in which text information is comprehensively represented from multiple granularities, and an \emph{ADR} is developed to refine images from both local and global perspectives.
\item Extensive experiments including quantitative and qualitative evaluations show the superiority and rationality of our proposed method. 
Specially, the causality study demonstrates~\shortname~as an interpretable model.
\end{itemize}

\section{Related Work}
\label{s:related work}

Due to the great potential in broad applications, text-to-image synthesis though challenging yet is arousing extensive research attention.
Earlier methods have achieved progress on this task due to the emergence of deep generative models~\cite{mansimov2015generating,gregor2015draw,nguyen2017plug,van2016conditional,reed2017parallel}.

Thanks to the advancement of GAN, recent approaches further improve the generation quality and have shown promising results on text-to-image synthesis.
Reed et al.~\cite{reed2016generative} first developed a simple and effective GAN architecture that enabled compelling text-to-image synthesis.
Nevertheless, the size of the image was only as small as $64\times64$.
To this end, StackGAN~\cite{zhang2017stackgan} was proposed to generate higher resolution images with two stages.
They initially sketched primary shape and colors, and then re-read the text to produce a photo-realistic image.
With the aim of discarding stacking architectures, Tao et al.~\cite{tao2020df} proposed DF-GAN to directly synthesize images without extra networks.
However, these works only took sentence-level features into consideration, which lacked fine-grained text understanding.
As a consequence, fine-grained details are often missing in the generated images.

To address this issue,
plenty of work has pushed text-to-image synthesis a large step forward
by utilizing word-level features at the refinement stage to enhance image details.
Among them, AttnGAN~\cite{xu2018attngan} played an important role. It utilized attention mechanism to repeatedly select important words at different steps for image refinement, which brought text-to-image synthesis research to a new height.
Zhu et al.~\cite{zhu2019dm} proposed DM-GAN that substituted memory network for attention mechanism to dynamically pick important words at the refinement stage.
To improve semantic consistency in text-to-image synthesis, Qiao et al.~\cite{qiao2019mirrorgan} proposed MirroGAN by semantically aligning the re-description of the generated image with the given text description. 
To explore the semantic correlation between different yet related sentences,
RiFeGAN~\cite{cheng2020rifegan} exploited an attention-based caption matching model to select and refine the compatible candidate captions from prior knowledge.
Yang et al.~\cite{yang2021multi} proposed MA-GAN to reduce the variation between their generated images with similar captions and enhance the reliability of the generated results.

\begin{figure*}
	\centering
	\includegraphics[height=55mm, width=165mm]{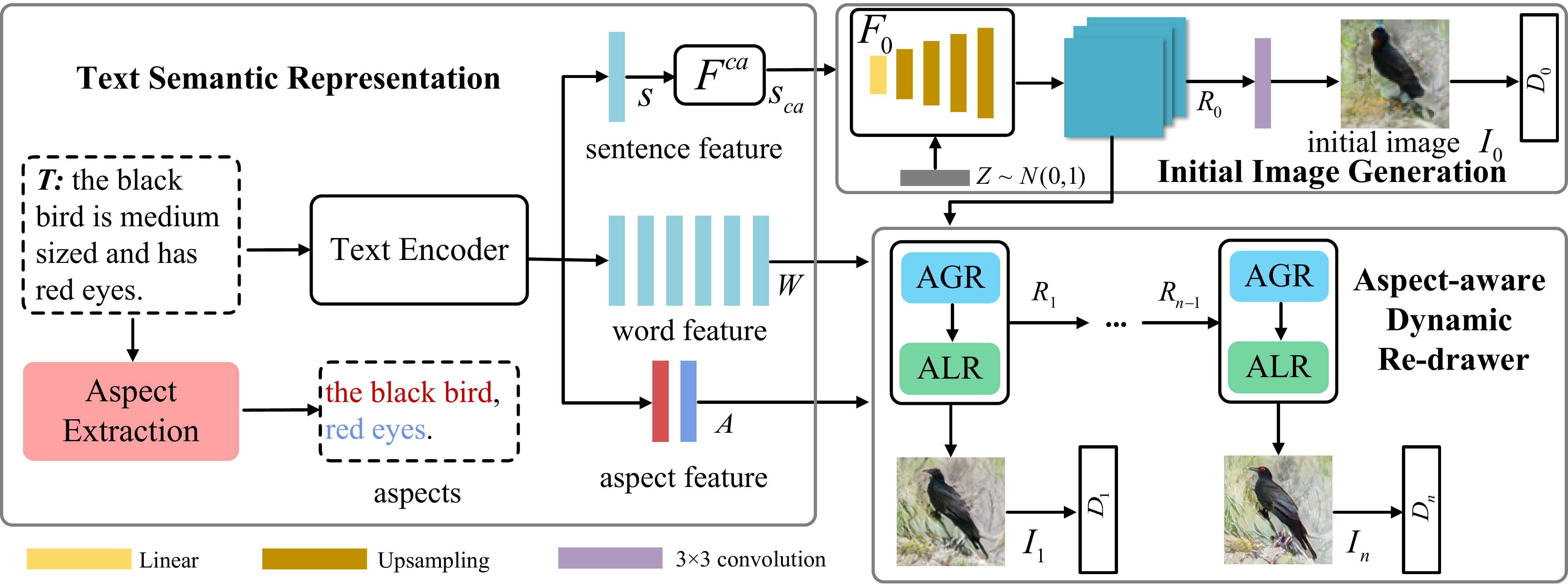}
	% 	\vspace{-2mm}
	\caption{Overall framework of \shortname.
	}
	\label{f:model}
		\vspace{-5mm}
\end{figure*}

With demands for various applications as well as the emergence of new data sets,
other compelling text-to-image research is also developed based on GAN.
In~\cite{johnson2018image,hong2018inferring,sun2019image,li2019object,huang2019realistic},
image generation was studied for datasets with multiple objects.
For example, Huang et al.~\cite{huang2019realistic} introduced an additional set of natural attentions between object-grid regions and word phrases.
However, extra bounding box information of each object must be required as labels.
To address the problem of food image synthesis with recipes, large efforts were made in~\cite{zhu2020cookgan,papadopoulos2019make,zhu2019r2gan}.
Other recent work also has achieved impressive results in person image synthesis~\cite{zhou2019text,ma2017pose}.
Given the reference images and text descriptions,
they could manipulate the visual appearance of a person.
Aiming at text-guided multi-modal face generation and manipulation, TediGAN as well as a face image dataset were proposed by Xia et al.~\cite{xia2020tedigan}.
To model the text and image tokens as a single stream of data,
Ramesh et al.~\cite{ramesh2021zero} proposed DALL$\cdot$E to train a transformer~\cite{vaswani2017attention} autoregressively.
Based on sufficient data and scale,
DALL$\cdot$E achieved comparable
results with other domain-specific models.

However, most of the aforementioned methods only considered the sentence-level and word-level features for text utilization. 
They ignored the great potential of aspect information contained in the sentences, which is very helpful for image refinement (\textit{e.g.}, the example in Figure~\ref{f:examp1e}).
To this end, in this paper, we argue that the aspect information should gain more attention and propose a novel text-to-image synthesis method to employ aspect-level features for local region refinement in a dynamic manner.

\section{Dynamic Aspect-aware GAN (DAE-GAN)}
\label{s:model}

As illustrated in Figure~\ref{f:model}, our proposed~\shortname~embodies three main components: 1) \textit{Text Semantic Representation}: extracting text semantic representations from multiple granularities, \textit{i.e.}, sentence-level, word-level, as well as aspect-level; 
2) \textit{Initial Image Generation}: generating a low-resolution image with sentence-level text features and a random noise vector; 
and 3) \textit{Aspect-aware Dynamic Re-drawer}: refining the initial image in a dynamic manner from both global and local perspectives, which is also the main focus in this paper.

\subsection{Text Semantic Representation}
Comprehensive understanding of text semantics plays a vital role in text-to-image synthesis.
Previous methods mainly extract text features from sentence-level and word-level.
However, they overlook aspect information contained in the text, 
which refers to several words rather than a word that depicts a particular part or feature of something, \textit{e.g.}, `\texttt{red eyes}' in `\texttt{the black bird is medium sized and has red eyes}'.
The granularity of aspect-level information is between those of sentence-level and word-level information.
It could be helpful for the refinement of image details and should gain more attention.
As shown in Figure~\ref{f:model}, 
we represent text features from multiple granularities, \textit{i.e.}, sentence-level, word-level, and aspect-level.
We use a Long Short-Term Memory~(LSTM) network to extract the semantic embedding of text description~$T$, which is formulated as follows:
\begin{equation}
\bm{s}, \bm{W} = \text{LSTM}(T),
\end{equation}
where $T=\{T_j|j=0, 1, ..., l-1\}$ consists of $l$ words.
$\bm{W}=\{\bm{W}_j|j=0, 1, ..., l-1\}\in\mathbb{R}^{l\times{d_w}}$ represents  word-level features that are obtained from the hidden state of LSTM at each time step.
Here, $d_w$ means the dimension of text embedding.
$\bm{s}\in\mathbb{R}^{d_w}$ stands for sentence-level semantic features from the last hidden state of LSTM.

We further employ the Conditioning Augmentation~(CA)~\cite{zhang2017stackgan} to augment the training data and avoid overfitting by resampling the input sentence vector from an independent Gaussian distribution.
Specifically, we enhance sentence features with CA, which is represented as follows:
\begin{equation}
\bm{s}_{ca} = F^{ca}(\bm{s}),
\end{equation}
where $F^{ca}(\bm{\cdot})$ stands for the CA function, and $\bm{s}_{ca}$ is the augmented sentence semantic representations with CA.

As mentioned before, aspect information is very crucial for the details of generated images. 
However, since the focus and description of different sentences are different, it is hard to identify and extract the proper aspect information for each sentence. 
To this end, we employ the syntactic structures to tackle this problem. 
Specifically, we first adopt NLTK to make the POS Tagging for each sentence. 
Then, we manually design different rules to extract aspect information for different datasets.
After that, we can obtain the aspect information $\{\bm{asp}_i|i=0,1,...,n-1\}$. 
Next, we use an LSTM to integrate this information and extract the aspect-level features, which is formulated as follows:

\begin{equation}
\bm{A} = \text{LSTM}(\{\bm{asp}_i|i=0,1,...,n-1\}), 
\end{equation}
where $\bm{A}$ denotes the aspect-level feature representation for text description and $n$ is the the number of extracted aspects.

\subsection{Initial Image Generation}
Following the common practice, we first generate a low-resolution image at the initial stage.
As illustrated in Figure~\ref{f:model},
we utilize the augmented sentence embeddings $\bm{s}_{ca}$ and a random noise vector $\bm{z}$ to generate an initial image $I_0$.
$\bm{z}\sim N(0,1)$ is sampled from a normal distribution.
Mathematically, we use $\bm{R}_0$ to denote the corresponding image features at the initial stage:
\begin{equation}
\bm{R}_{0} = F_0(\bm{s}_{ca}, \bm{z}),
\end{equation}
where $F_0$ is the image generator at the initial generation stage.
As depicted in Figure~\ref{f:model}, it consists of one fully connected layer and four upsampling layers.

\subsection{Aspect-aware Dynamic Re-drawer}
To the best of our knowledge, we are the first to introduce aspect information contained in the given sentence into text-to-image synthesis.
Therefore, how to integrate the aspect information into image refinement stage
is the main challenge that we should tackle.
Inspired by human learning behaviors, in this work, we develop a novel \emph{Aspect-aware Dynamic Re-drawer~(ADR)} to refine images with the consideration of aspect information in the sentence. 
Specifically, we design a novel \emph{Attended Global Refinement~(AGR)} module to employ fine-grained word-level features for global refinement, and a novel \emph{Aspect-aware Local Refinement~(ALR)} module to utilize aspect-level features for local enhancement. 
By alternately applying these two components in a dynamic way, we are able to refine image details from both global and local perspectives. 
In the following part, we will take the $i^{th}$ refinement operation for generated image as an example to introduce the technical details of \emph{AGR} and \emph{ALR}.

\begin{figure}
	\centering
	\includegraphics[height=74mm, width=79mm]{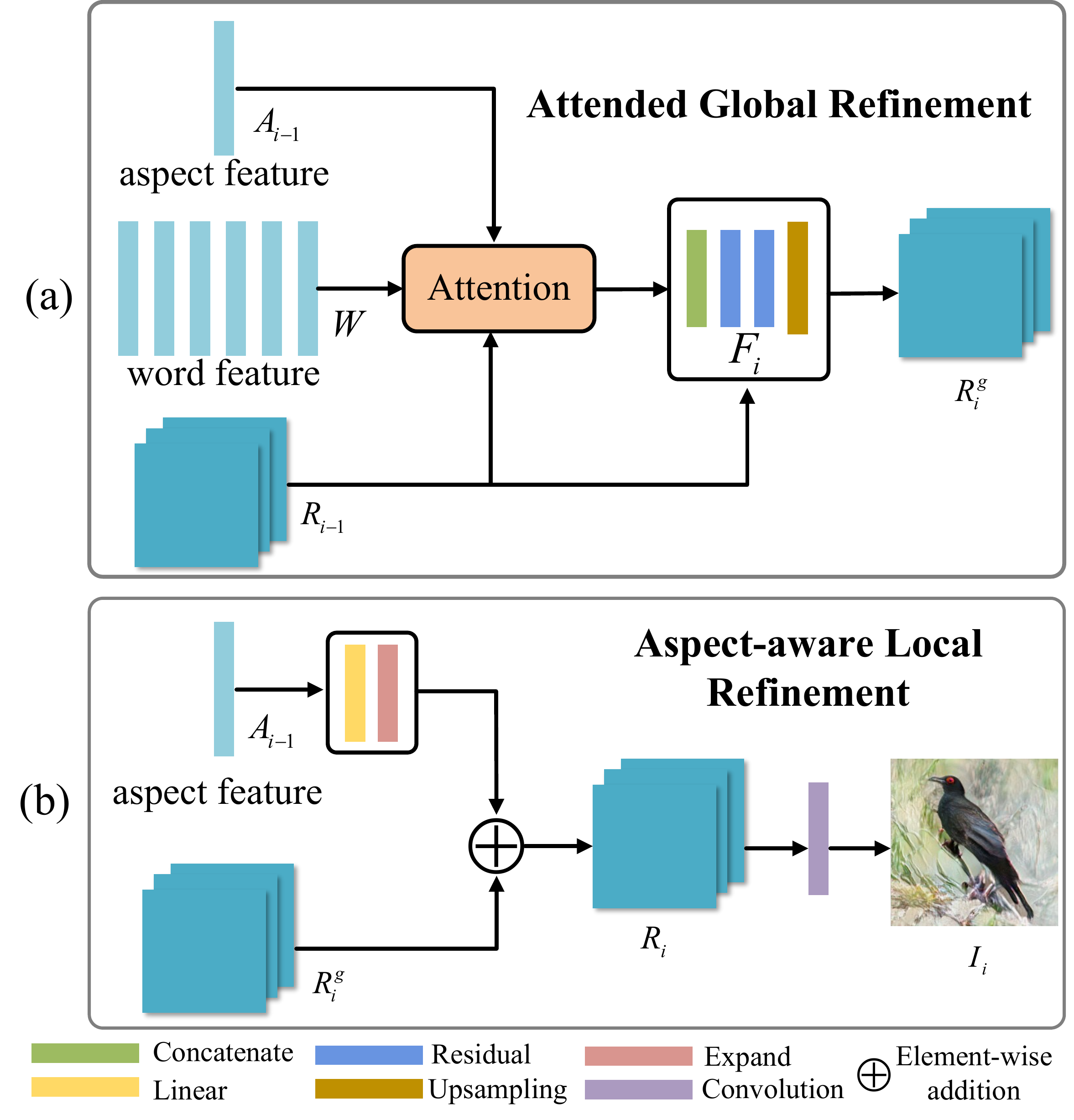}
	% 	\vspace{-2mm}
	\caption{Architecture of \emph{AGR} and \emph{ALR}.}
	\label{f:ADR}
	\vspace{-6mm}
\end{figure}

\subsubsection{Attended Global Refinement}
To synthesize a photo-realistic and semantic consistent image, it is necessary to further globally refine the image with fine-grained features.
Therefore,  
\emph{AGR} is developed for global refinement based on the initial image.

Specifically,
we use word-level text features to help the refinement process by taking into account the contribution of each word.
Current works mainly update the word-level features by employing image features from the previous step to select important words with attention mechanism~\cite{ruan2020context}.
Differently, we make a further step to integrate both image features and  aspect-level features to update and enhance word-level features, as depicted in Figure~\ref{f:ADR}~(a).
This process can be mathematically formulated as follows:
\begin{equation}
\begin{split}
\bm{R}_{i}^{g} &= F_i(\bm{R}_{i-1}, \bm{W}_{i}^{g}), i = 1, 2, ..., n,\\
\bm{W}_{i}^{g} &= \sum_{j=0}^{l-1}(\bm{W}_j\bm{U})\alpha_{i,j},\\ 
\alpha_{i,j} &= \text{softmax}((\bm{W}_j\bm{U}+\bm{A}_{i-1}\bm{V})\bm{R}_{i-1}),
\end{split}
\end{equation}
where $\bm{R}_{i}^{g}\in\mathbb{R}^{d_r\times{N_i}}$ represents the image features enriched globally with image features $\bm{R}_{i-1}\in\mathbb{R}^{d_r\times{N_{i-1}}}$ and attended word-level features.
$N_i$ is the size of $R_i^g$ at the $i^{th}$ step.
$F_i(\bm{\cdot,~\bm{\cdot}})$ denotes the image feature transformer.
$\bm{W}_{i}^{g}\in\mathbb{R}^{d_r\times{N_{i-1}}}$ means the attended global features.
$\alpha_{i,j}$ stands for the attention weight scores.
$\bm{U}\in\mathbb{R}^{d_w\times{d_r}}$ and $\bm{V}\in\mathbb{R}^{d_w\times{d_r}}$ are perception layers to convert word embedding $\bm{W}$ and aspect embedding $\bm{A}$ into an underlying common semantic space of visual features.

\subsubsection{Aspect-aware Local Refinement}
In the previous part, we have introduced how to utilize word-level features to refine images from the global perspective.
However, the enhancement of some specific image local details has not been completed yet.
As mentioned above, aspects contained in the text description could be significant for synthesizing the corresponding local image details.
To this end, as shown in Figure~\ref{f:ADR}~(b),
\emph{ALR} is developed to refine images from a local perspective with aspect-level features.

Technically, we combine the aspect features $\bm{A}_{i-1}$ with globally refined image features $\bm{R}_i^g$ by element-wise addition as follows:
\begin{equation}
\bm{R}_{i} = \bm{R}_i^g + [\bm{A}_{i-1}\bm{V}]\otimes{N_i}, i = 1, 2, ..., n,
\end{equation}
where operation $A_i\otimes{N_i} = [A_i;A_i;...;A_i]$ means repeatedly concatenating $A_i$ for $N_i$ times. 
To synthesize a photo-realistic image,
we finally introduce a $3\times{3}$ convolution filter to transform the refined image feature $\bm{R}_i$ into image $I_i$ at the $i^{th}$ refinement operation in \emph{ADR}. 
In summary,
\emph{AGR} and \emph{ALR} are alternately applied.
Meanwhile, aspect-level features are dynamically added at each refinement step in \emph{ADR}.

\subsection{Objective Function}

To generate a photo-realistic image and ensure the semantic consistency between text description and the corresponding image simultaneously, we carefully design the loss function. 
During each step, the generator $G$~(\textit{e.g.}, \emph{ADR}) and the discriminator $D$ are trained in an alternative fashion.
To start with common practice, the objective loss function of each generator at each step is defined as follows:
\begin{equation}
\begin{aligned}
L_{G_i} = -\frac{1}{2}[&\underbrace{\mathbb{E}_{I_i\sim{p_{G_i}}}\log{D_i(I_i)}}_{\text{unconditional~ loss}}+\underbrace{\mathbb{E}_{I_i\sim{p_{G_i}}}\log{D_i(I_i,T)}}_{\text{conditional~loss}}],
\end{aligned}
\end{equation}
where the first unconditional loss term is derived from the discriminator in distinguishing between real and fake images.
The second term is a conditional loss to make the synthesized image match the input sentence.

Traditionally, the conditional loss term consists of sentence-image and word-image pairs.
Different from previous works,
we introduce aspect information through the generation process.
To ensure the generated images truly contain local fine-grained details that match the corresponding aspects, we also include an aspect-image matching pair in the conditional loss as follows:
\begin{equation}
D(I,T) = D(I,\bm{s})^{\beta_1}\cdot D(I,\bm{W})^{\beta_2}\cdot D(I, \bm{A})^{\beta_3}, 
\end{equation}
where $D(I,\bm{s})$, $D(I,\bm{W})$, and $D(I, \bm{A})$ calculate the matching degrees between the image and the sentence, word, and aspect, respectively.

Following~\cite{zhu2019dm,xu2018attngan}, we further utilize the DAMSM loss~\cite{xu2018attngan} to compute the matching degree between images and text descriptions, mathematically denoted as $L_{\text{DAMSM}}$.
And the CA loss is defined as the Kullback-Leibler divergence between the standard Gaussian distribution and the Gaussian distribution of training text, \textit{i.e.},
\begin{equation}
L_{CA} = D_{KL}(\mathcal{N}(\mu(\textbf{s}),{\sum}(\textbf{s}))||\mathcal{N}(0, I)).
\end{equation}

The final objective function of the generator networks is composed of the aforementioned three terms:
\begin{equation}
L_G = \sum_{i}L_{G_i}+\lambda_1L_{CA}+\lambda_2L_{DAMSM}.
\end{equation}

For adversarial learning, each discriminator $D_i$ is trained to precisely identify the input image as real or fake by minimizing the cross-entropy loss.
The adversarial loss for each discriminator $D_i$ is defined as:
\begin{equation}
\begin{aligned}
&L_{D_i} = -\frac{1}{2}[\underbrace{\mathbb{E}_{I_i^{GT}\sim{p_{GT}}}\log{D_i(I_i^{GT})}+\mathbb{E}_{I_i\sim{p_{G_i}}}\log{D_i(I_i)}}_{\text{unconditional~loss}} \\ &+\underbrace{\mathbb{E}_{I_i^{GT}\sim{p_{GT}}}\log{D}_i(I_i^{GT},T)
+\mathbb{E}_{I_i\sim{p_{G_i}}}\log{D_i(I_i,T)}}_{\text{conditional~loss}}],
\end{aligned}
\end{equation}
where the unconditional loss is responsible for distinguishing synthesized images from  real ones and the conditional term determines whether the image matches the input sentence.
$I_i^{GT}$ is sampled from the real image distribution $p_{GT}$ at the $i^{th}$ step.
The final objective function of the discriminator networks is $L_{D} = \sum_i L_{D_i}$.

\section{Experiment}
\label{s:experiment}

In this section, we will first introduce the experiment setup. 
Next, we will evaluate~\shortname~on two publicly available and well-studied datasets.
Then, visualization study as well as causality analysis will be discussed to show the effectiveness and interpretability of \shortname. 

\subsection{Experiment Setup}

\textbf{Datasets}.
To demonstrate the capability of our proposed method, we conduct extensive experiments on the CUB-200~\cite{wah2011caltech} and COCO~\cite{lin2014microsoft} datasets,
following previous text-to-image synthesis works~\cite{xu2018attngan,zhu2019dm,qiao2019mirrorgan,zhang2017stackgan}.
The CUB-200 dataset contains $200$ bird categories with $8,855$ training images and $2,933$ test images.
Each image in CUB-200 has 10 text captions.
For the COCO dataset, it consists of a training set with $80k$ images and a test set with $40k$ images.
There are 5 captions for each image in COCO.

\textbf{Evaluation Metrics}.
Following~\cite{xu2018attngan,zhu2019dm}, for better comparison, we quantitatively measure the performance of~\shortname~in terms of Inception Score~(IS)~\cite{salimans2016improved}, Fr\'echet Inception Distance~(FID)~\cite{heusel2017gans}, and R-precision~\cite{xu2018attngan}.

We obtain IS by employing a pre-trained Inception-v3 network~\cite{szegedy2016rethinking} to compute the KL-divergence between the conditional class distribution and the marginal class distribution.
A large IS indicates generated images have a high diversity for all classes,
and each of them could be clearly recognized as a specific class rather than an ambiguous one.

FID computes the Fr\'echet distance between the synthetic and real-world images based on the feature map output from the pre-trained Inception\_v3 network.
Lower FID score means a closer distance between the generated image distribution and real image distribution and therefore implies the model is capable of synthesizing photo-realistic images.

R-precision is used to evaluate the semantic consistency between the synthetic image and the given text description.
Similarly, we calculate the cosine distance between the global image vector and 100 candidate global sentence vectors to measure the image-text semantic similarity.
The lower R-precision means better semantic consistency between synthesized images and given text descriptions. 

\textbf{Implementation Details}.~\footnote{https://github.com/hiarsal/DAE-GAN}
For aspect rules,
each \textbf{(\underline{adjective}, \underline{noun})} pair is an aspect that describes an object or scene.
For COCO that contains layout and location, if a \textbf{preposition} is before the pair that expresses a relative spacial relationship, we will also add it.
Consistent with~\cite{xu2018attngan,zhu2019dm},
we adopt Inception-v3~\cite{szegedy2016rethinking} pre-trained on ImageNet~\cite{russakovsky2015imagenet} as image encoder, and use pre-trained LSTM~\cite{xu2018attngan} as text encoder.
The size of the initial generated low-resolution image~(\textit{i.e.}, $N_0$) is set to $64\times{64}$.
The finally synthesized high-resolution image at the last step has the size~($N_n$) of $256\times{256}$.
During intermediate steps, all the image size~($N_i$) is fixed to $128\times128$. 
Empirically, we set $d_w = 256$ and $d_r = 64$ to be the dimensions of text and image feature vectors, respectively.
For other related hyperparameters, we set $(\beta_1,~\beta_2,~\beta_3)=(1,~1,~0.2)$.
For CUB-200, we set $(\lambda_1,~\lambda_2,~n)=(1,~5,~2)$, and for COCO, $(\lambda_1,~\lambda_2,~n)=(1,~50,~3)$.
During training,
we use the Adam optimizer with the learning rate of $0.0002$ to train the networks on 8 NVIDIA Tesla V100 GPUs in parallel with the batch size of 32 on each one.
\shortname~is trained with 600 epochs and 120 epochs on CUB-200 and COCO respectively. 

\begin{table}
	\caption{Inception Score~(higher is better) on different models.}
	\label{t:IS}
	\vspace{1mm}
    \scalebox{0.82}{
		\begin{tabular}{lcc} 
			\hline
			Model & CUB-200 & COCO \\ 
			\hline
			(1)~GAN-INT-CLS~\cite{reed2016generative} & 2.88$\pm$0.04 & 7.88$\pm$0.07 \\
			
			(2)~StackGAN~\cite{zhang2017stackgan} & 3.70$\pm$0.04 & 8.45$\pm$0.03 \\
			\hline
			(3)~AttnGAN~\cite{xu2018attngan} & 4.36$\pm$0.03 & 25.89$\pm$0.47 \\
			
			(4)~MirroGAN~\cite{qiao2019mirrorgan} & 4.54$\pm$0.17 & 26.47$\pm$0.41 \\
			
			(5)~Huang et al.~\cite{huang2019realistic} &- &26.92$\pm$0.52 \\
			
			(6)~DM-GAN~\cite{zhu2019dm} & 4.75$\pm$0.07 & 30.49$\pm$0.57 \\
			
			(7)~LostGAN~\cite{sun2019image} & - & 13.8$\pm$0.4 \\
			
			(8)~MA-GAN~\cite{yang2021multi} & 4.76$\pm$0.09 & - \\
			
			(9)~KT-GAN~\cite{tan2020kt} & 4.85$\pm$0.04 & \underline{31.67$\pm$0.36} \\
			
			(10)~DF-GAN~\cite{tao2020df} & \underline{4.86$\pm$0.04} & - \\
			
			(11)~RiFe-GAN~\cite{cheng2020rifegan} & \textbf{5.23$\pm$0.09} & - \\
			\hline
			(12)~\shortname & 4.42$\pm$0.04 & \textbf{35.08$\pm$1.16}\\
			\hline
		\end{tabular}
            }
	\centering
	\vspace{-6mm}
\end{table}

\subsection{Quantitative Results}

\begin{figure*}
	\centering
	\includegraphics[height=72mm, width=176mm]{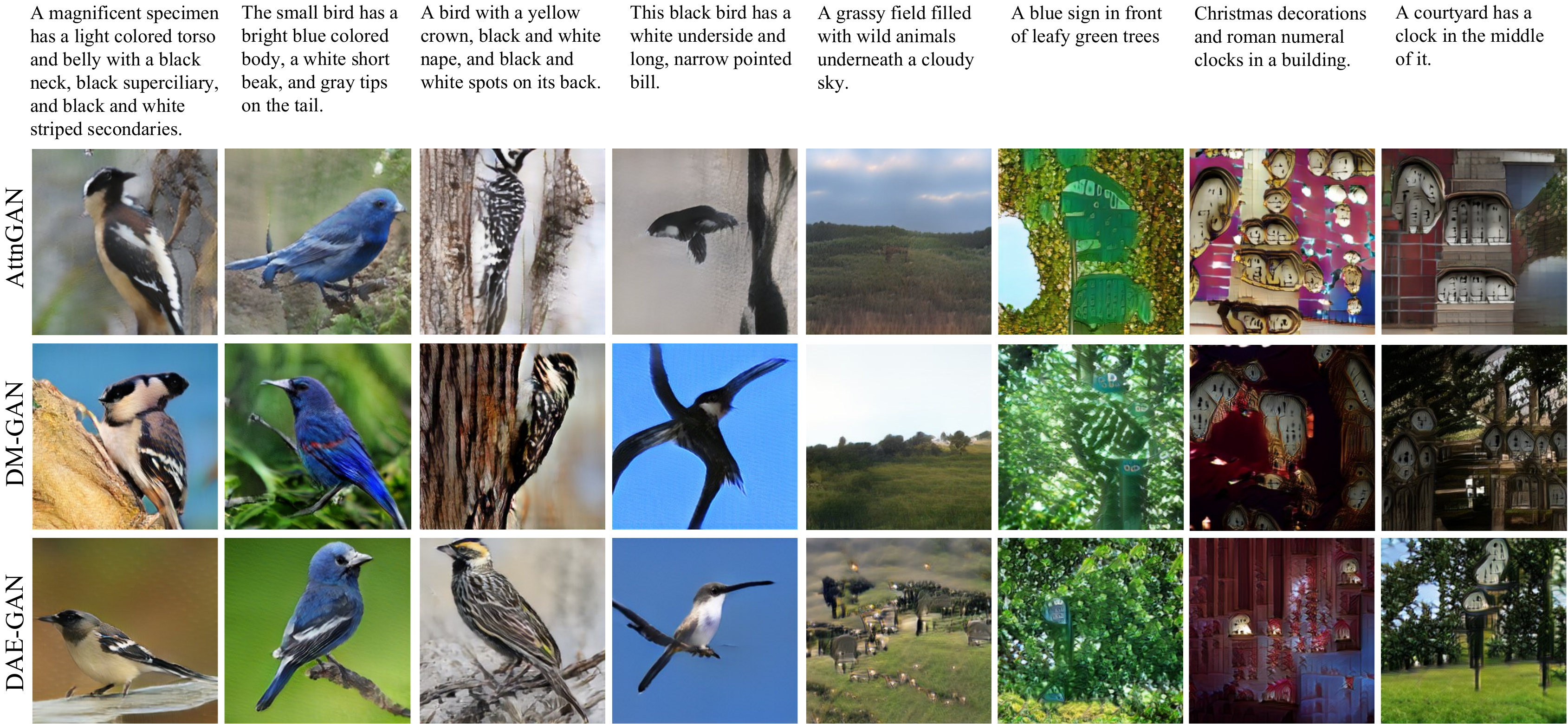}
	% 	\vspace{-2mm}
	\caption{Example results for text-to-image synthesis by AttnGAN~\cite{xu2018attngan}, DM-GAN~\cite{zhu2019dm} and our proposed~\shortname~on CUB-200~(the left four ones) and COCO~(the right four ones).
	}
	\label{f:visual_quality}
		\vspace{-5mm}
\end{figure*}

\textbf{Performance on CUB-200.}
We compare our methods with state-of-the-art methods on CUB-200.
The overall results are summarized in Table~\ref{t:IS},~\ref{t:FID}, and \ref{t:r-precision}.

It is clear that our proposed~\shortname~achieves highly comparable performance, especially for FID and R-precision scores that measure photo realism and semantic consistency respectively.
Specifically,~\shortname~first learns comprehensive text semantics from multiple granularities, \textit{i.e.}, sentence-level, word-level, as well as aspect-level.
This is one of the reasons that~\shortname~could improve FID and R-precision scores against other baselines by a large margin.
Moreover, \emph{ADR}, the core component of~\shortname, is developed to refine images by alternately applying \emph{AGR} and \emph{ALR} in a dynamic manner, in which 
\emph{AGR} enhances images from the global perspective with word-level features and  \emph{ALR} refines images from the local perspective with aspect-level features.
This is another vital reason to make our synthesized image more photo-realistic and keep semantic consistency between text and image.
    
CUB-200 is a dataset full of description details.
Therefore, models with comprehensive text semantic understanding achieve much better results than coarse-grained ones.
For example, GAN-INT-CLS and StackGAN only leverage sentence-level features as input. 
Based on them, AttnGAN and DM-GAN employ word-level features to refine images and achieve higher performance.
RiFeGAN is particularly designed for datasets with fine-grained visual details, \textit{e.g.}, CUB-200.
It synthesizes images from multiple captions,
which on the one hand leads to very high IS due to more caption details,
on the other hand causes low R-precision score.
Differently, our~\shortname~can synthesize images with high FID and R-precision scores only from one given caption.
This is largely achieved by comprehensive representation and utilization of text information, including sentence-level, word-level as well as aspect-level.

\begin{table}
	%	\vspace{-1mm}
	\caption{FID score~(lower is better) on different models.}
	\vspace{1mm}
	\scalebox{0.82}{
		\begin{tabular}{lcc} 
			\hline
			Model & CUB-200 & COCO \\ 
			\hline
			(1)~AttnGAN~\cite{xu2018attngan} & 23.98 & 35.49 \\
			
			(2)~Huang et al.~\cite{huang2019realistic} & - & 34.52 \\
			
			(3)~MA-GAN~\cite{yang2021multi} & 21.66 & - \\
			
			(4)~DM-GAN~\cite{zhu2019dm} & \underline{16.09} & 32.64  \\
			
			(5)~KT-GAN~\cite{tan2020kt} & 17.32 & 30.73 \\
			
			(6)~LostGAN~\cite{sun2019image} & - & 29.65 \\
			
			(7)~DF-GAN~\cite{tao2020df} & 19.24 & \underline{28.92} \\
			\hline
			(8)~\shortname & \textbf{15.19} & \textbf{28.12}\\ 
			\hline
		\end{tabular}
	}
	\centering
	\label{t:FID}
	\vspace{-2mm}
\end{table}

\begin{table}
    \caption{R-precision~(\%)~(higher is better) on different models.}
	\vspace{1mm}
	\scalebox{0.82}{
		\begin{tabular}{lcc} 
			\hline
			Model & CUB-200 & COCO \\ 
			\hline
			(1)~AttnGAN~\cite{xu2018attngan} & 67.82$\pm$4.43 & 72.31$\pm$0.91\\
			
			(2)~MirroGAN~\cite{qiao2019mirrorgan} & 57.67 & 74.52 \\
		
			(3)~DM-GAN~\cite{zhu2019dm} & \underline{72.31$\pm$0.91} & 88.56$\pm$0.28  \\
			
			(4)~Huang et al.~\cite{huang2019realistic} & - & \underline{89.69$\pm$4.34} \\
			
			(5)~RiFeGAN~\cite{cheng2020rifegan} & 23.8$\pm$1.5 & - \\
			\hline
			(6)~\shortname & \textbf{85.45$\pm$0.57} & \textbf{92.61$\pm$0.50} \\
			\hline
		\end{tabular}
	}
	\centering
	\label{t:r-precision}
	\vspace{-6mm}
\end{table}

\textbf{Performance on COCO.}
We also evaluate our method on COCO that has multiple objects, complex layout and simple details.
The relative results are reported in Table~\ref{t:IS}, \ref{t:FID} and \ref{t:r-precision}.
We also list the observations as follows:

\shortname~still achieves the best quantitative performance against baseline methods with regard to IS, FID and R-precision.
The results demonstrate that~\shortname~is also capable of synthesizing well semantic consistent images with multi-object and complex layout.
Comprehensive understanding of conditional text description and the newly proposed refinement paradigm \emph{ADR} make it possible for~\shortname~to refine different objects with dynamically provided aspect information.
It is also the main reason that~\shortname~can generalize well on different datasets.

\subsection{Qualitative Results}
To evaluate the visual quality of generated images, we show some subjective comparisons among AttnGAN~\cite{xu2018attngan}, DM-GAN~\cite{zhu2019dm} and our proposed~\shortname~in Figure~\ref{f:visual_quality}.

In CUB-200, we can obtain that~\shortname~generates better results.
For example,
when synthesizing a bird with the detail of long narrow bill~($4^{th}$ column), only~\shortname~achieves this goal.
Also in the $1^{st}$ and $3^{rd}$ columns, only~\shortname~synthesizes photo-realistic and semantic consistency images.
The reason is that~\shortname~obtains comprehensive text representations, especially aspect-level features.
Moreover, \emph{ALR} is developed to dynamically enhance image details with aspect information.

In COCO dataset, we can also observe that images generated by~\shortname~are more vivid and realistic.
Taking examples in $6^{th}$ and $7^{th}$ columns of Figure~\ref{f:visual_quality}, AttnGAN and DM-GAN often generate one object multiple times and the spatial distribution is also chaotic~\cite{xu2018attngan,zhu2019dm}, while~\shortname~could address the problems well.
By alternately applying \emph{AGR} and \emph{ALR},~\shortname~will not only enhance local details, but also refine images from a global perspective.
This mechanism allows~\shortname~to avoid getting stuck in a few most important words like other methods.

\subsection{Ablation Study}

\begin{table}
    \caption{Ablation performance on COCO about IS, FID and R-precision~(\%).}
	\vspace{1mm}
	\scalebox{0.82}{
		\begin{tabular}{lccc} 
			\hline
			Model & IS & FID & R-precision \\ 
			\hline
			\shortname~(w/o AGR) & 2.93$\pm$0.03 &149.79 & 2.34$\pm$0.26\\
			
			\shortname~(w/o ALR) & 31.07$\pm$0.70 & 32.93 & 90.24$\pm$0.39 \\
		    \hline
		    \shortname~(w/o asp in AGR) & 34.70$\pm$0.64 & 28.60 & 92.28$\pm$0.46 \\
			\hline
			\shortname & \textbf{35.08$\pm$1.16} & \textbf{28.12} & \textbf{92.61$\pm$0.50}  \\
			\hline
		\end{tabular}
	}
    % \vspace{2mm}
	\centering
	\label{t:ablation study}
	\vspace{-6mm}
\end{table}

The overall experiments have proved the superiority of our proposed~\shortname.
However, which component is really important for performance improvement is still unclear.
Therefore, we perform an ablation study on COCO to verify the effectiveness of each part in \emph{ADR}, including \emph{AGR} and \emph{ALR}. 
Corresponding results are illustrated in Table~\ref{t:ablation study}.
According to the results, we can observe varying degrees of model performance decline when removing \emph{AGR} and \emph{ALR} separately from~\shortname.
Recalling the aspect, since ALR is dependent on aspect information, we further remove aspect from AGR.
The model performance also declines.
The ablation study demonstrates comprehensive utilization of text information is helpful for image synthesis.
\emph{AGR} and \emph{ALR} could employ the information well for image refinement.

\begin{figure*}
	\centering
	\includegraphics[height=37mm, width=158mm]{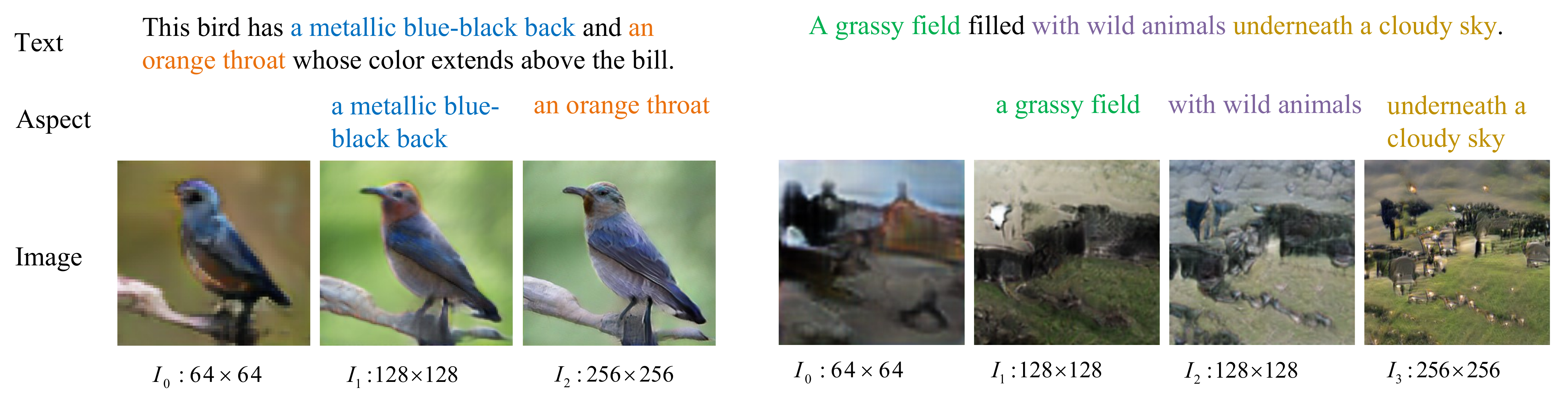}
	% 	\vspace{-2mm}
	\caption{Text-to-image synthesis visualization of different generation steps.
	}
	\label{f:generation_process}
		\vspace{-4mm}
\end{figure*}

\begin{figure}
	\centering
	\includegraphics[height=60mm, width=84mm]{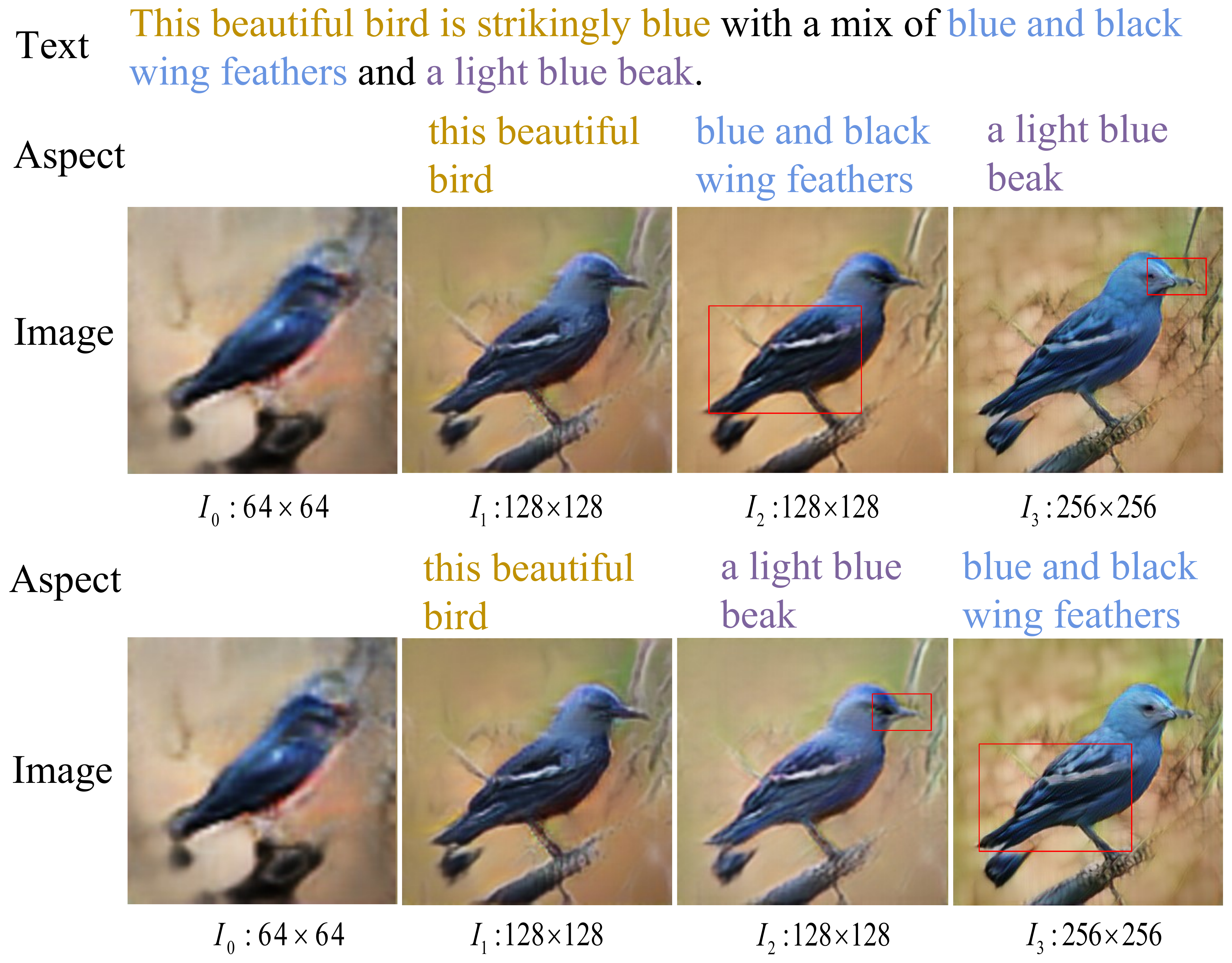}
	% 	\vspace{-2mm}
	\caption{Cause-and-effect study of~\shortname. We exchange the input order of the last two aspect information to explore how the images generated in the related steps will correspondingly change with the variation of aspect input orders.}
	\label{f:attribute_order}
	\vspace{-5mm}
\end{figure}

\subsection{Causality Interpretation}
\textbf{Visualization of Generation Process.}
To evaluate model rationality and interpretability, we study the synthesis process in Figure~\ref{f:generation_process}.
In the left example, \shortname~initially generates a low-resolution image~(\textit{i.e.}, $I_0$) with the whole sentence.
Then, based on $I_0$, \emph{ADR} further employs fine-grained information~(\textit{i.e.}, word-level and aspect-level features) to refine the image.
Specifically, in image $I_1$, \emph{ADR} pours its attention to the refinement respecting the aspect information of `\texttt{a metallic blue-black back}' and improves the image size to $128\times128$.
Finally in $I_2$, \emph{ADR} focuses on the refinement of aspect `\texttt{an orange throat}' and further improves the resolution to $256\times256$.
Moreover, we can observe that 
in $I_1$, almost the bird head is in orange, while in $I_2$, only the throat is correctly drawn in orange and the image looks more vivid.
In the right example, we present a visualization study in COCO with three aspects.
We could observe that the images are also well generated under the corresponding guide of aspects `\texttt{a grassy field}', `\texttt{with wild animals}' and `\texttt{underneath a cloudy sky}'.

\textbf{Influence of Aspect Order.}
As aspect information is placed in a significant position in our model, and it is dynamically utilized to refine local details,
we are curious whether the order of aspect input will affect the generation results.
Thus, an example case is studied in Figure~\ref{f:attribute_order}.
To be specific, 
the given text description has three aspect features as shown in Figure~\ref{f:attribute_order}.
We exchange the input order of the last two aspect features to explore how the images generated at the related steps will correspondingly change with the variation of aspect input orders.
Meanwhile, we will not change the input of sentence-level features and word-level features.
Taking the look at the examples of upper part, when refining the aspect of `\texttt{blue and black wing feathers}', we could find the bird feathers are vividly drawn in blue and black.
Then in $I_3$, it is obvious that the color of bird beak is changed from light black in $I_2$ to light blue.
In the examples of bottom part, \emph{ADR} focuses on the aspect `\texttt{a light blue beak}' in $I_2$.
Visually compared with $I_1$, the bird beak color turned light blue from light black.
Then in $I_3$, the feathers are refined more vividly in blue and black than the ones in $I_2$.

The experimental results again fully confirmed the importance of aspect information to image refinement. 
Meanwhile,~\shortname~can make full use of the aspect information to achieve image refinement in a dynamic manner.
Furtherly, these examples also illustrate that our proposed \shortname~has good interpretation.

\section{Conclusion}
\label{s:conclusion}

In this paper, we argued that aspect information contained in the text is quite helpful for image generation and should gain more attention.
Then, we developed a novel~\shortname~to make full use of aspect information for text-to-image synthesis.
To be specific,
we utilized text information from multiple granularities, including sentence-level, word-level and aspect-level.
Moreover, a new generation paradigm \emph{ADR} was developed to refine initial images, in which a novel \emph{AGR} was proposed to refine images from the global perspective and a novel \emph{ALR} was designed to enhance image details from the local perspective.
By employing these two components dynamically, our proposed \shortname~had the ability to leverage aspect information to refine the details of the generated image, which is critical for image realism and semantic consistency.
Extensive experiments demonstrated the superiority and rationality of our proposed method.
In the future, we will pay a special focus on exploring a way to improve semantic consistency with self-supervised contrastive learning.
\section*{Acknowledgment}

This research was partially supported by grants from the National Natural Science Foundation of China (Grants No. 61727809, 61922073, 62006066, and U20A20229), and the Open Project Program of the National Laboratory of Pattern Recognition (NLPR).

{\small
\bibliographystyle{ieee_fullname}
\bibliography{egbib}

\begin{thebibliography}{10}\itemsep=-1pt

\bibitem{brandt1973differential}
Th Brandt, Johannes Dichgans, and Ellen Koenig.
\newblock Differential effects of central versus peripheral vision on
  egocentric and exocentric motion perception.
\newblock {\em Experimental brain research}, 16(5):476--491, 1973.

\bibitem{chen2018text2shape}
Kevin Chen, Christopher~B Choy, Manolis Savva, Angel~X Chang, Thomas
  Funkhouser, and Silvio Savarese.
\newblock Text2shape: Generating shapes from natural language by learning joint
  embeddings.
\newblock In {\em Asian Conference on Computer Vision}, pages 100--116.
  Springer, 2018.

\bibitem{chen2020relation}
Zhuang Chen and Tieyun Qian.
\newblock Relation-aware collaborative learning for unified aspect-based
  sentiment analysis.
\newblock In {\em Proceedings of the 58th Annual Meeting of the Association for
  Computational Linguistics}, pages 3685--3694, 2020.

\bibitem{cheng2020rifegan}
Jun Cheng, Fuxiang Wu, Yanling Tian, Lei Wang, and Dapeng Tao.
\newblock Rifegan: Rich feature generation for text-to-image synthesis from
  prior knowledge.
\newblock In {\em Proceedings of the IEEE/CVF Conference on Computer Vision and
  Pattern Recognition}, pages 10911--10920, 2020.

\bibitem{goodfellow2014generative}
Ian Goodfellow, Jean Pouget-Abadie, Mehdi Mirza, Bing Xu, David Warde-Farley,
  Sherjil Ozair, Aaron Courville, and Yoshua Bengio.
\newblock Generative adversarial nets.
\newblock In {\em Advances in neural information processing systems}, pages
  2672--2680, 2014.

\bibitem{gregor2015draw}
Karol Gregor, Ivo Danihelka, Alex Graves, Danilo~Jimenez Rezende, and Daan
  Wierstra.
\newblock Draw: A recurrent neural network for image generation.
\newblock {\em arXiv preprint arXiv:1502.04623}, 2015.

\bibitem{heusel2017gans}
Martin Heusel, Hubert Ramsauer, Thomas Unterthiner, Bernhard Nessler, and Sepp
  Hochreiter.
\newblock Gans trained by a two time-scale update rule converge to a local nash
  equilibrium.
\newblock In {\em Advances in neural information processing systems}, pages
  6626--6637, 2017.

\bibitem{hong2018inferring}
Seunghoon Hong, Dingdong Yang, Jongwook Choi, and Honglak Lee.
\newblock Inferring semantic layout for hierarchical text-to-image synthesis.
\newblock In {\em Proceedings of the IEEE Conference on Computer Vision and
  Pattern Recognition}, pages 7986--7994, 2018.

\bibitem{huang2019realistic}
Wanming Huang, Richard~Yi Da~Xu, and Ian Oppermann.
\newblock Realistic image generation using region-phrase attention.
\newblock In {\em Asian Conference on Machine Learning}, pages 284--299. PMLR,
  2019.

\bibitem{johnson2018image}
Justin Johnson, Agrim Gupta, and Li Fei-Fei.
\newblock Image generation from scene graphs.
\newblock In {\em Proceedings of the IEEE conference on computer vision and
  pattern recognition}, pages 1219--1228, 2018.

\bibitem{li2019object}
Wenbo Li, Pengchuan Zhang, Lei Zhang, Qiuyuan Huang, Xiaodong He, Siwei Lyu,
  and Jianfeng Gao.
\newblock Object-driven text-to-image synthesis via adversarial training.
\newblock In {\em Proceedings of the IEEE Conference on Computer Vision and
  Pattern Recognition}, pages 12174--12182, 2019.

\bibitem{lin2014microsoft}
Tsung-Yi Lin, Michael Maire, Serge Belongie, James Hays, Pietro Perona, Deva
  Ramanan, Piotr Doll{\'a}r, and C~Lawrence Zitnick.
\newblock Microsoft coco: Common objects in context.
\newblock In {\em European conference on computer vision}, pages 740--755.
  Springer, 2014.

\bibitem{ma2017pose}
Liqian Ma, Xu Jia, Qianru Sun, Bernt Schiele, Tinne Tuytelaars, and Luc
  Van~Gool.
\newblock Pose guided person image generation.
\newblock In {\em Advances in neural information processing systems}, pages
  406--416, 2017.

\bibitem{mansimov2015generating}
Elman Mansimov, Emilio Parisotto, Jimmy~Lei Ba, and Ruslan Salakhutdinov.
\newblock Generating images from captions with attention.
\newblock {\em arXiv preprint arXiv:1511.02793}, 2015.

\bibitem{nguyen2017plug}
Anh Nguyen, Jeff Clune, Yoshua Bengio, Alexey Dosovitskiy, and Jason Yosinski.
\newblock Plug \& play generative networks: Conditional iterative generation of
  images in latent space.
\newblock In {\em Proceedings of the IEEE Conference on Computer Vision and
  Pattern Recognition}, pages 4467--4477, 2017.

\bibitem{papadopoulos2019make}
Dim~P Papadopoulos, Youssef Tamaazousti, Ferda Ofli, Ingmar Weber, and Antonio
  Torralba.
\newblock How to make a pizza: Learning a compositional layer-based gan model.
\newblock In {\em Proceedings of the IEEE Conference on Computer Vision and
  Pattern Recognition}, pages 8002--8011, 2019.

\bibitem{peng2020knowing}
Haiyun Peng, Lu Xu, Lidong Bing, Fei Huang, Wei Lu, and Luo Si.
\newblock Knowing what, how and why: A near complete solution for aspect-based
  sentiment analysis.
\newblock In {\em Proceedings of the AAAI Conference on Artificial
  Intelligence}, volume~34, pages 8600--8607, 2020.

\bibitem{qiao2019mirrorgan}
Tingting Qiao, Jing Zhang, Duanqing Xu, and Dacheng Tao.
\newblock Mirrorgan: Learning text-to-image generation by redescription.
\newblock In {\em Proceedings of the IEEE Conference on Computer Vision and
  Pattern Recognition}, pages 1505--1514, 2019.

\bibitem{ramesh2021zero}
Aditya Ramesh, Mikhail Pavlov, Gabriel Goh, Scott Gray, Chelsea Voss, Alec
  Radford, Mark Chen, and Ilya Sutskever.
\newblock Zero-shot text-to-image generation.
\newblock {\em arXiv preprint arXiv:2102.12092}, 2021.

\bibitem{reed2016generative}
Scott Reed, Zeynep Akata, Xinchen Yan, Lajanugen Logeswaran, Bernt Schiele, and
  Honglak Lee.
\newblock Generative adversarial text to image synthesis.
\newblock In {\em 33rd International Conference on Machine Learning}, pages
  1060--1069, 2016.

\bibitem{reed2017parallel}
Scott~E Reed, A{\"a}ron van~den Oord, Nal Kalchbrenner, Sergio~Gomez
  Colmenarejo, Ziyu Wang, Yutian Chen, Dan Belov, and Nando de Freitas.
\newblock Parallel multiscale autoregressive density estimation.
\newblock In {\em ICML}, 2017.

\bibitem{ruan2020context}
Shulan Ruan, Kun Zhang, Yijun Wang, Hanqing Tao, Weidong He, Guangyi Lv, and
  Enhong Chen.
\newblock Context-aware generation-based net for multi-label visual emotion
  recognition.
\newblock In {\em 2020 IEEE International Conference on Multimedia and Expo
  (ICME)}, pages 1--6. IEEE, 2020.

\bibitem{russakovsky2015imagenet}
Olga Russakovsky, Jia Deng, Hao Su, Jonathan Krause, Sanjeev Satheesh, Sean Ma,
  Zhiheng Huang, Andrej Karpathy, Aditya Khosla, Michael Bernstein, et~al.
\newblock Imagenet large scale visual recognition challenge.
\newblock {\em International journal of computer vision}, 115(3):211--252,
  2015.

\bibitem{salimans2016improved}
Tim Salimans, Ian Goodfellow, Wojciech Zaremba, Vicki Cheung, Alec Radford, and
  Xi Chen.
\newblock Improved techniques for training gans.
\newblock {\em Advances in neural information processing systems},
  29:2234--2242, 2016.

\bibitem{strasburger2011peripheral}
Hans Strasburger, Ingo Rentschler, and Martin J{\"u}ttner.
\newblock Peripheral vision and pattern recognition: A review.
\newblock {\em Journal of vision}, 11(5):13--13, 2011.

\bibitem{sun2019image}
Wei Sun and Tianfu Wu.
\newblock Image synthesis from reconfigurable layout and style.
\newblock In {\em Proceedings of the IEEE International Conference on Computer
  Vision}, pages 10531--10540, 2019.

\bibitem{szegedy2016rethinking}
Christian Szegedy, Vincent Vanhoucke, Sergey Ioffe, Jon Shlens, and Zbigniew
  Wojna.
\newblock Rethinking the inception architecture for computer vision.
\newblock In {\em Proceedings of the IEEE conference on computer vision and
  pattern recognition}, pages 2818--2826, 2016.

\bibitem{tan2020kt}
Hongchen Tan, Xiuping Liu, Meng Liu, Baocai Yin, and Xin Li.
\newblock Kt-gan: Knowledge-transfer generative adversarial network for
  text-to-image synthesis.
\newblock {\em IEEE Transactions on Image Processing}, 30:1275--1290, 2020.

\bibitem{tao2020df}
Ming Tao, Hao Tang, Songsong Wu, Nicu Sebe, Fei Wu, and Xiao-Yuan Jing.
\newblock Df-gan: Deep fusion generative adversarial networks for text-to-image
  synthesis.
\newblock {\em arXiv preprint arXiv:2008.05865}, 2020.

\bibitem{van2016conditional}
Aaron Van~den Oord, Nal Kalchbrenner, Lasse Espeholt, Oriol Vinyals, Alex
  Graves, et~al.
\newblock Conditional image generation with pixelcnn decoders.
\newblock {\em Advances in neural information processing systems},
  29:4790--4798, 2016.

\bibitem{vaswani2017attention}
Ashish Vaswani, Noam Shazeer, Niki Parmar, Jakob Uszkoreit, Llion Jones,
  Aidan~N Gomez, Lukasz Kaiser, and Illia Polosukhin.
\newblock Attention is all you need.
\newblock {\em arXiv preprint arXiv:1706.03762}, 2017.

\bibitem{wah2011caltech}
Catherine Wah, Steve Branson, Peter Welinder, Pietro Perona, and Serge
  Belongie.
\newblock The caltech-ucsd birds-200-2011 dataset.
\newblock 2011.

\bibitem{wang2016attention}
Yequan Wang, Minlie Huang, Xiaoyan Zhu, and Li Zhao.
\newblock Attention-based lstm for aspect-level sentiment classification.
\newblock In {\em Proceedings of the 2016 conference on empirical methods in
  natural language processing}, pages 606--615, 2016.

\bibitem{warren1992role}
William~H Warren and Kenneth~J Kurtz.
\newblock The role of central and peripheral vision in perceiving the direction
  of self-motion.
\newblock {\em Perception \& Psychophysics}, 51(5):443--454, 1992.

\bibitem{xia2020tedigan}
Weihao Xia, Yujiu Yang, Jing-Hao Xue, and Baoyuan Wu.
\newblock Tedigan: Text-guided diverse image generation and manipulation.
\newblock {\em arXiv preprint arXiv:2012.03308}, 2020.

\bibitem{xu2018attngan}
Tao Xu, Pengchuan Zhang, Qiuyuan Huang, Han Zhang, Zhe Gan, Xiaolei Huang, and
  Xiaodong He.
\newblock Attngan: Fine-grained text to image generation with attentional
  generative adversarial networks.
\newblock In {\em Proceedings of the IEEE conference on computer vision and
  pattern recognition}, pages 1316--1324, 2018.

\bibitem{yang2021multi}
Yanhua Yang, Lei Wang, De Xie, Cheng Deng, and Dacheng Tao.
\newblock Multi-sentence auxiliary adversarial networks for fine-grained
  text-to-image synthesis.
\newblock {\em IEEE Transactions on Image Processing}, 30:2798--2809, 2021.

\bibitem{zhang2017stackgan}
Han Zhang, Tao Xu, Hongsheng Li, Shaoting Zhang, Xiaogang Wang, Xiaolei Huang,
  and Dimitris~N Metaxas.
\newblock Stackgan: Text to photo-realistic image synthesis with stacked
  generative adversarial networks.
\newblock In {\em Proceedings of the IEEE international conference on computer
  vision}, pages 5907--5915, 2017.

\bibitem{zhang2018stackgan++}
Han Zhang, Tao Xu, Hongsheng Li, Shaoting Zhang, Xiaogang Wang, Xiaolei Huang,
  and Dimitris~N Metaxas.
\newblock Stackgan++: Realistic image synthesis with stacked generative
  adversarial networks.
\newblock {\em IEEE transactions on pattern analysis and machine intelligence},
  41(8):1947--1962, 2018.

\bibitem{zheng2020replicate}
Yaowei Zheng, Richong Zhang, Samuel Mensah, and Yongyi Mao.
\newblock Replicate, walk, and stop on syntax: An effective neural network
  model for aspect-level sentiment classification.
\newblock In {\em Proceedings of the AAAI Conference on Artificial
  Intelligence}, volume~34, pages 9685--9692, 2020.

\bibitem{zhi2017pixelbrush}
Jiale Zhi.
\newblock Pixelbrush: Art generation from text with gans.
\newblock In {\em Cl. Proj. Stanford CS231N Convolutional Neural Networks Vis.
  Recognition, Sprint 2017}, page 256, 2017.

\bibitem{zhou2019text}
Xingran Zhou, Siyu Huang, Bin Li, Yingming Li, Jiachen Li, and Zhongfei Zhang.
\newblock Text guided person image synthesis.
\newblock In {\em Proceedings of the IEEE Conference on Computer Vision and
  Pattern Recognition}, pages 3663--3672, 2019.

\bibitem{zhu2020cookgan}
Bin Zhu and Chong-Wah Ngo.
\newblock Cookgan: Causality based text-to-image synthesis.
\newblock In {\em Proceedings of the IEEE/CVF Conference on Computer Vision and
  Pattern Recognition}, pages 5519--5527, 2020.

\bibitem{zhu2019r2gan}
Bin Zhu, Chong-Wah Ngo, Jingjing Chen, and Yanbin Hao.
\newblock R2gan: Cross-modal recipe retrieval with generative adversarial
  network.
\newblock In {\em Proceedings of the IEEE Conference on Computer Vision and
  Pattern Recognition}, pages 11477--11486, 2019.

\bibitem{zhu2019dm}
Minfeng Zhu, Pingbo Pan, Wei Chen, and Yi Yang.
\newblock Dm-gan: Dynamic memory generative adversarial networks for
  text-to-image synthesis.
\newblock In {\em Proceedings of the IEEE Conference on Computer Vision and
  Pattern Recognition}, pages 5802--5810, 2019.

\end{thebibliography}
}

\end{document}